\documentclass[letterpaper]{article}

\usepackage{aaai2026}

\makeatletter
\gdef\copyright@on{} 
\makeatother

\usepackage{times}
\usepackage{helvet}
\usepackage{courier}
\usepackage{amsmath,amssymb,amsfonts}
\usepackage{graphicx}
\usepackage{booktabs}
\usepackage{multirow}
\usepackage{tabularx}

\usepackage[hyphens]{url}
\usepackage[numbers]{natbib}
\setcitestyle{numbers,square,comma,sort&compress}

\newcommand{\CBL}{\textsc{ChaosBench-Logic}}
\newcommand{\FOL}{first-order logic}

\newcommand{\Chaotic}{\mathsf{Chaotic}}
\newcommand{\Deterministic}{\mathsf{Deterministic}}
\newcommand{\PosLyap}{\mathsf{PosLyap}}
\newcommand{\Sensitive}{\mathsf{Sensitive}}
\newcommand{\StrangeAttr}{\mathsf{StrangeAttr}}
\newcommand{\PointUnpredictable}{\mathsf{PointUnpredictable}}
\newcommand{\StatPredictable}{\mathsf{StatPredictable}}
\newcommand{\QuasiPeriodic}{\mathsf{QuasiPeriodic}}
\newcommand{\Random}{\mathsf{Random}}
\newcommand{\FixedPointAttr}{\mathsf{FixedPointAttr}}
\newcommand{\Periodic}{\mathsf{Periodic}}

\title{ChaosBench-Logic: A Benchmark for Logical and Symbolic Reasoning\\on Chaotic Dynamical Systems}

\author{
  Noel Thomas
}
\affiliations{
  Mohamed bin Zayed University of Artificial Intelligence (MBZUAI)\\
  noel.thomas@mbzuai.ac.ae
}

\begin{document}
\maketitle

\begin{abstract}
Large language models (LLMs) excel at natural language tasks but remain
brittle in domains requiring precise logical and symbolic reasoning.
Chaotic dynamical systems provide an especially demanding test because chaos
is deterministic yet frequently conflated with randomness, complexity, or
mere nonlinearity.
This paper introduces \CBL, a benchmark that evaluates LLM reasoning across 30 diverse
dynamical systems using a unified \FOL\ ontology.
Each system is annotated with truth assignments for 11 semantic predicates,
and 621 questions are generated across seven reasoning categories, including
multi-hop implications, cross-system analogies, counterfactual reasoning,
bias probes, and multi-turn dialogues.
Metrics are defined for logical accuracy, implication consistency, dialogue
coherence, and contradiction, and an open-source evaluation pipeline is released.
Initial experiments show that frontier LLMs such as GPT-4, Claude~3.5 Sonnet,
Gemini~2.5 Flash, and the open-source LLaMA-3~70B achieve 91--94\% per-item
accuracy, yet still score 0\% on compositional items and exhibit fragile
global coherence: dialogue-level accuracy ranges from 53.1\% (GPT-4 CoT)
to 75.5\% (LLaMA-3 zeroshot).
\CBL provides a rigorous testbed for diagnosing these failures and a foundation
for developing neuro-symbolic approaches that improve scientific reasoning in LLMs.
\end{abstract}

\begin{links}
  \link{\mbox{Code (GitHub)}}{https://github.com/11NOel11/ChaosBench-Logic}
  \link{\mbox{Dataset (Hugging Face)}}{https://huggingface.co/datasets/11NOel11/ChaosBench-Logic}
\end{links}

\section{Introduction}

Large language models transform natural language processing,
demonstrating strong abilities in translation, summarisation, question
answering, and code synthesis. Despite this progress, they remain
surprisingly brittle when confronted with tasks requiring precise logical or
symbolic reasoning. In scientific domains such as dynamical systems and
chaos theory, accurate reasoning is vital: one must distinguish whether a
system is chaotic, periodic, quasi-periodic, or random; infer implications
of qualitative properties; and reason about parameter perturbations and
bifurcations. Errors in such reasoning can lead to incorrect scientific
conclusions and undermine trust in LLM-based scientific tools.

Chaotic dynamics provide a stringent test of reasoning because ``chaos'' is
often misunderstood even by experts. Chaos is deterministic yet exhibits
sensitive dependence on initial conditions, positive Lyapunov exponents,
and complex attractor geometry~\cite{strogatz2015nonlinear,ott2002chaos,devaney1989intro}.
It is \emph{not} equivalent to randomness, nonlinearity, or complexity;
confusing these concepts causes systematic misinterpretations of physical
systems. Existing LLM benchmarks do not adequately test this mix of domain
semantics and formal reasoning depth. This paper therefore develops a
benchmark that systematically evaluates and analyses LLM reasoning about
chaotic and non-chaotic systems under formal logical constraints.

\subsection{Contributions}
\label{sec:contributions}

This paper makes the following key contributions:

\begin{enumerate}
  \item \textbf{\CBL\ Benchmark.} A benchmark of 621 reasoning questions over
        30 dynamical systems spanning ODEs, maps, PDEs, neuronal and chemical
        models, and stochastic processes~\cite{thomas2026chaosbenchlogic,thomas2026chaosbenchlogic_hf}.
  \item \textbf{\FOL\ Ontology.} A unified logical vocabulary (11 predicates)
        and a compact axiom system capturing widely accepted implications
        between chaos, determinism, Lyapunov behavior, randomness,
        attractor types, and predictability.
  \item \textbf{Seven Task Families.} Atomic QA, multi-hop implications,
        cross-system analogies and non-analogies, counterfactual reasoning,
        bias probes, multi-turn dialogues, and hard compositional questions.
  \item \textbf{Metrics and Pipeline.} Reproducible evaluation for accuracy,
        implication consistency, dialogue coherence, contradiction detection,
        and axiom-violation counts.
  \item \textbf{Empirical Diagnosis.} Frontier and open-source LLMs achieve
        high per-item accuracy (91--94\%) but fail on compositional items
        (0\%) and show fragile multi-turn coherence.
\end{enumerate}

\section{Related Work}
\label{sec:related}

\paragraph{Logical and Symbolic Reasoning in LLMs.}
Recent work assesses LLMs on deductive and symbolic tasks, including
BIG-Bench~\cite{srivastava2022bigbench}, ReClor~\cite{yu2020reclor},
LogiQA~\cite{liu2020logiqa}, and related logic benchmarks. These datasets
reveal persistent brittleness in multi-step inference, contradiction
avoidance, entailment, and reasoning under explicit constraints. Several
efforts introduce structured prompting (e.g., chain-of-thought~\cite{wei2022chainofthought}) or tool-augmented reasoning, yet models still hallucinate intermediate steps or fail to maintain coherence across turns.
\CBL extends this line of inquiry to a scientific domain where reasoning must
stay consistent with formal definitions of chaos-related semantics.

\paragraph{Scientific QA and Domain Reasoning.}
Benchmarks such as SciBench~\cite{wang2023scibench} and MATH~\cite{hendrycks2021math}
evaluate quantitative or symbolic scientific problem solving, highlighting
limitations in equation manipulation and multi-step reasoning. However, none
target dynamical systems theory, nor do they ground questions in a shared
logical ontology. \CBL differs by anchoring all questions in a unified \FOL\
structure and scoring models for logical coherence rather than only
final-answer plausibility.

\paragraph{Chaos and Machine Learning.}
Prior ML-for-chaos work primarily evaluates forecasting or attractor
reconstruction (e.g., reservoir computing and deep learning baselines)
rather than formal reasoning about qualitative regimes~\cite{pathak2018reservoir,gilpin2021chaos}.
\CBL complements these benchmarks by focusing on symbolic and logical
competence: distinguishing chaos from randomness and enforcing consistent
implication structure across diverse systems.

\paragraph{Bias and Misconception Diagnostics.}
Evaluations such as TruthfulQA~\cite{lin2022truthfulqa} examine systematic
model errors and truthfulness failures. Yet domain-specific misconceptions
(e.g., chaos--randomness confusion, nonlinearity--chaos conflation) remain
largely unexplored. \CBL introduces explicit bias-probe families tailored to
chaos theory semantics.

\section{Methods}
\label{sec:methods}

This section formalizes \CBL, describes the ontology and axiom system,
details dataset construction, and defines evaluation metrics.

\subsection{Problem Definition}

This paper considers reasoning over a collection of dynamical systems
$\mathcal{S}=\{s_1,\dots, s_{30}\}$ including continuous-time flows,
discrete-time maps, PDEs, neuronal oscillators, chemical reaction models,
and stochastic processes.

Each system $s \in \mathcal{S}$ is associated with semantic properties
expressed as unary logical predicates from a fixed vocabulary $\mathcal{P}$.
Given a natural-language query about a system $s$, a model must predict
the logically correct answer according to the system's ground-truth property
assignment and a global axiom system.

\subsection{Logical Predicate Vocabulary}

\CBL defines 11 unary predicates, each mapping a system $s$ to a boolean:
\begin{itemize}
  \item $\Chaotic(s)$
  \item $\Deterministic(s)$
  \item $\PosLyap(s)$ (positive largest Lyapunov exponent)
  \item $\Sensitive(s)$
  \item $\StrangeAttr(s)$
  \item $\PointUnpredictable(s)$
  \item $\StatPredictable(s)$
  \item $\QuasiPeriodic(s)$
  \item $\Random(s)$ (true randomness / stochasticity)
  \item $\FixedPointAttr(s)$
  \item $\Periodic(s)$
\end{itemize}

\subsection{Predicate Semantics and Annotation Protocol}
\label{sec:predicate-semantics}

To make the benchmark auditable and to reduce ambiguity, each
predicate is treated as a semantic label over a named system (or canonical regime)
rather than as an invitation for models to interpret definitions ad hoc.
In particular, the benchmark is designed so that a system can appear random
in its trajectories and still be labeled deterministic, reflecting the core
distinction between stochasticity and sensitive determinism.

This paper uses the following annotation principles:
\begin{itemize}
  \item \textbf{Determinism vs.\ randomness.} $\Deterministic(s)$ indicates
  that the system evolution is fully specified by its state and update rule.
  $\Random(s)$ is reserved for systems with inherent stochasticity (e.g.,
  explicit noise terms or stochastic differential equations).
  \item \textbf{Chaos as a regime label.} $\Chaotic(s)$ indicates a canonical
  parameter regime widely treated as chaotic in the literature for that named
  system. Borderline regimes are avoided and disputed cases are removed during curation.
  \item \textbf{Predictability predicates.} $\PointUnpredictable(s)$ captures
  practical loss of long-horizon point forecasts even under perfect modeling,
  while $\StatPredictable(s)$ reflects that coarse statistical structure can
  remain predictable or stable over time (e.g., invariant measures).
  \item \textbf{Attractor-type predicates.} $\Periodic(s)$, $\QuasiPeriodic(s)$,
  and $\FixedPointAttr(s)$ are mutually exclusive indicators of common
  non-chaotic behavior classes, used to generate non-analogies and ``do not
  overgeneralize'' items.
\end{itemize}

These principles make \CBL closer to a controlled reasoning benchmark than a
definition-discovery task: models are evaluated on logical consistency under
a fixed ontology and ground truth.

\subsection{Global \FOL\ Axiom System}
\label{sec:axioms}

\CBL specifies a set $\Phi$ of global axioms encoding widely used
implications in dynamical systems~\cite{strogatz2015nonlinear,ott2002chaos,devaney1989intro}.
Each implication is written explicitly:

\paragraph{Chaos axioms.}
\begin{align}
\forall s:\; \Chaotic(s) &\Rightarrow \Deterministic(s), \\
\forall s:\; \Chaotic(s) &\Rightarrow \PosLyap(s), \\
\forall s:\; \Chaotic(s) &\Rightarrow \Sensitive(s), \\
\forall s:\; \Chaotic(s) &\Rightarrow \PointUnpredictable(s), \\
\forall s:\; \Chaotic(s) &\Rightarrow \StatPredictable(s), \\
\forall s:\; \Chaotic(s) &\Rightarrow \neg \Random(s).
\end{align}

\paragraph{Lyapunov \& predictability.}
\begin{align}
\forall s:\; \PosLyap(s) &\Rightarrow \Sensitive(s), \\
\forall s:\; \PosLyap(s) &\Rightarrow \PointUnpredictable(s).
\end{align}

\paragraph{Attractor-type implications.}
\begin{align}
\forall s:\; \StrangeAttr(s) &\Rightarrow \Chaotic(s), \\
\forall s:\; \Periodic(s) &\Rightarrow \neg \Chaotic(s), \\
\forall s:\; \QuasiPeriodic(s) &\Rightarrow \neg \Chaotic(s), \\
\forall s:\; \FixedPointAttr(s) &\Rightarrow \neg \Chaotic(s).
\end{align}

\paragraph{Randomness vs determinism.}
\begin{align}
\forall s:\; \Random(s) &\Rightarrow \neg \Deterministic(s).
\end{align}

\paragraph{Design choice: no converse rules.}
Crucially, reverse implications are not assumed (e.g.,
$\Sensitive \Rightarrow \Chaotic$). This avoids overspecification and forces
models to respect directionality: many scientific misconceptions arise from
treating one-way implications as equivalences.

\subsection{Ground-Truth Closure and Answer Derivation}
\label{sec:closure}

The benchmark ground truth for each system is a truth assignment for the
11 predicates that is consistent with $\Phi$. For implication questions,
the correct answer is computed by applying forward chaining under $\Phi$.

Let $\mathcal{A}_s$ denote the set of all literals known true/false for a
system $s$. The logical closure under $\Phi$ is defined as:
\begin{equation}
\begin{aligned}
\mathrm{Cl}_\Phi(\mathcal{A}_s)
&= \mathrm{lfp}\Bigl(\mathcal{A}_s \cup
\{\,\ell \;:\; (\ell_1 \wedge \cdots \wedge \ell_k \Rightarrow \ell)\in \Phi, \\
&\qquad\qquad\ \ \ell_1,\dots,\ell_k \in \mathcal{A}_s \}\Bigr),
\end{aligned}
\end{equation}
where $\mathrm{lfp}$ is the least fixed point reached by repeatedly applying
all applicable implications. Because the system annotations satisfy $\Phi$,
the closure remains consistent. For model outputs, closure and
axiom checks expose contradictions and implication violations.

\subsection{System Coverage}

\CBL includes 30 systems across five broad classes. Table~\ref{tab:systems}
provides an overview.

\begin{table}[t]
\centering
\caption{Overview of dynamical systems included in \CBL. Each system is
annotated with 11 semantic predicates and participates in \FOL\ reasoning.}
\begin{tabularx}{\columnwidth}{lXl}
\toprule
Class & Examples & Chaos? \\
\midrule
Chaotic ODEs &
Lorenz-63, Lorenz-84, Lorenz-96, R\"ossler, Chen, HR neuron &
yes \\
Non-chaotic ODEs &
Simple harmonic oscillator (SHM), Van der Pol, Brusselator, FitzHugh--Nagumo &
no \\
Maps &
Logistic, H\'enon, Ikeda, Baker, Arnold Cat, Standard Map, Circle Map &
mixed \\
PDEs &
Kuramoto--Sivashinsky (chaotic), Sine--Gordon (integrable) &
mixed \\
Stochastic &
Ornstein--Uhlenbeck &
random, non-chaotic \\
\bottomrule
\end{tabularx}
\label{tab:systems}
\end{table}

\begin{samepage}
This paper includes canonical systems spanning flows and maps (e.g., Lorenz and
R\"ossler systems, H\'enon and logistic maps) as well as PDE and stochastic
processes \cite{lorenz1963dnf,rossler1976equation,henon1976map,may1976simple,
sivashinsky1977nonlinear,kuramoto1978diffusion,ornstein1930brownian}.
\end{samepage}

\subsection{Dataset Construction Pipeline}

The benchmark is produced via a four-stage pipeline:

\paragraph{(1) System encoding.}
Each system is encoded in a structured JSON file containing equations or
update rules, canonical parameter regimes, short natural-language summaries,
and predicate truth assignments. This representation supports both human
inspection and programmatic question generation.

\paragraph{(2) Template-based generation with semantic constraints.}
This paper designs seven task families with $70+$ templates. Templates are not
free-form: each includes explicit semantic constraints (e.g., ``only apply
this template to systems with $\Random(s)=\text{TRUE}$'') to avoid ill-posed
questions. Lexical surface forms and question structure are also varied to
reduce shallow cue exploitation.

\paragraph{(3) Human curation for correctness and ambiguity removal.}
All candidates are manually checked for logical correctness under
$\mathrm{Cl}_\Phi(\cdot)$, linguistic clarity, and consistency with the
intended misconception being tested (for bias probes). Ambiguous items
(e.g., parameter regimes with disputed behavior) are removed.

\paragraph{(4) Final filtering and consistency checks.}
Questions that violate applicability rules or create logical self-conflicts
are automatically filtered, followed by a final manual pass. In addition to
basic sanity checks, each item is verified to be answerable from the system
labels and $\Phi$ alone (i.e., without numerical simulation).

\subsection{Question Format and Metadata}
\label{sec:format}

Each benchmark item includes a system identifier and a natural-language
question, plus structured metadata used in evaluation. The current release
includes fields such as \texttt{system\_id}, \texttt{family},
\texttt{template\_id}, and (for dialogues) \texttt{dialogue\_id} and
\texttt{turn\_index}. This makes it possible to compute aggregate metrics
per family and to audit failures at the level of a specific template or
system.

For compositional items, the metadata explicitly records which sub-skills are
required (e.g., implication chaining plus analogy rejection), enabling more
fine-grained analysis than a single global accuracy score.

\subsection{Reasoning Task Families}
\label{sec:families}

\CBL covers seven reasoning types:

\begin{enumerate}
  \item \textbf{Atomic Logical QA:} single-step predicate queries.
  \item \textbf{Multi-Hop Implication Reasoning:} forward-chaining under $\Phi$.
  \item \textbf{Cross-System Analogical / Non-Analogical Reasoning:} transfer and rejection.
  \item \textbf{Counterfactual Parameter Reasoning:} bifurcations and perturbations.
  \item \textbf{Bias Probes:} adversarial misconception tests.
  \item \textbf{Advanced Multi-Turn Dialogues:} cross-turn coherence and contradiction avoidance.
  \item \textbf{Hard Concept-Synthesis / Compositional:} multiple reasoning modes combined.
\end{enumerate}

Table~\ref{tab:family-purpose} summarizes what each family measures.

\begin{table}[t]
\centering
\small
\caption{What each task family tests and why it is non-trivial for LLMs.}
\begin{tabularx}{\columnwidth}{lX}
\toprule
Family & What it measures (typical failure) \\
\midrule
Atomic QA & Correctly recall system-level truth values (confident hallucination). \\
Multi-hop & Apply directionality of implications (converse fallacy). \\
Analogy & Transfer only the shared structure (surface-form analogy bias). \\
Counterfactual & Reason about regime change vs invariance (overgeneralize chaos). \\
Bias probes & Resist misleading cues (``chaos = random'', ``PDE = chaotic''). \\
Dialogues & Maintain consistent belief state over turns (belief drift). \\
Compositional & Integrate multiple constraints simultaneously (local-but-not-global reasoning). \\
\bottomrule
\end{tabularx}
\label{tab:family-purpose}
\end{table}

\subsection{Multi-Turn Dialogue Design}

Each dialogue is grouped under a \texttt{dialogue\_id}, with 3--6 turns.
Correctness requires per-turn accuracy and intra-dialogue coherence:
answers must not contradict earlier commitments under $\Phi$.
This targets persistent reasoning and belief-state stability.

A key design choice is that dialogue turns intentionally mix ``easy'' atomic
questions with implication-driven follow-ups. This stresses whether a model
can (i) answer a local question, (ii) remember its own earlier commitment,
and (iii) apply the same commitment consistently when queried from a different
angle later in the conversation.

\subsection{Bias Families}
\label{sec:bias-families}

This paper identifies six systematic reasoning biases:
\begin{enumerate}
  \item Chaos--Randomness Bias
  \item Chaos--Nonlinearity Bias
  \item Complexity--Chaos Bias
  \item Fractal--Chaos Bias
  \item PDE--Chaos Overgeneralisation
  \item Name/Analogy Bias (``Lorenz-like $\rightarrow$ chaotic'')
\end{enumerate}

These biases are encoded at the template level in the current release and can
be refined into per-item labels in future versions.

\subsection{Evaluation Metrics}
\label{sec:metrics}

This paper introduces metrics that separate local correctness from
global coherence.

\paragraph{Logical accuracy.}
Given $N_{\text{eval}}$ questions with valid normalized predictions:
\begin{equation}
\mathrm{Acc}
= \frac{1}{N_{\mathrm{eval}}}
  \sum_{i=1}^{N_{\mathrm{eval}}}
  \mathrm{I}\bigl(\hat{y}_i = y_i^*\bigr).
\end{equation}

\paragraph{Dialogue accuracy.}
For a dialogue $d$ with turns $t=1,\dots,T_d$:
\begin{equation}
\mathrm{DialogueAcc}(d)
= \mathrm{I}\bigl(\forall t,\; \hat{y}_{d,t} = y_{d,t}^*\bigr).
\end{equation}
This metric is strict by design: a single incoherent turn collapses a whole
dialogue to 0.

\paragraph{Contradiction \& axiom-violation checks.}
Beyond correctness, this work evaluates whether a model's set of commitments
is internally consistent. Contradictions (e.g., predicting both
$\Chaotic(s)$ and $\Random(s)$) are flagged and violations of implications in $\Phi$
are counted within a dialogue. These diagnostics explain how and why
dialogues fail even when per-turn answers look locally plausible.

\section{Experimental Setup and Baselines}
\label{sec:experiments}

\paragraph{Models.}
This paper evaluates representative frontier and open-source LLMs:
GPT-4 (OpenAI), Claude~3.5 Sonnet (Anthropic), Gemini~2.5 Flash (Google),
and LLaMA-3~70B (Meta)~\cite{openai2023gpt4,anthropic2024claude35,google2025gemini25flash,meta2024llama3}.
All models are queried through their official APIs or hosted endpoints.

\paragraph{Prompting.}
Zero-shot prompts request a final YES/NO decision.
For chain-of-thought (CoT) prompting, each model is asked to write reasoning
and end with \texttt{FINAL\_ANSWER: YES/NO}~\cite{wei2022chainofthought}.
The evaluation script extracts and normalizes the final label.

\paragraph{Hyperparameters and determinism.}
All models are run at temperature 0 (or nearest deterministic setting) to
reduce sampling variance and emphasize reasoning consistency failures.

\paragraph{Evaluation pipeline.}
Responses are normalized using a robust pattern matcher. If a response
contains multiple answers, the final explicit marker (CoT) or the
last YES/NO token (zero-shot) is preferred. This avoids over-crediting
models that hedge or self-correct mid-output.

\section{Results}
\label{sec:results}

This section presents empirical results on \CBL. All accuracies are computed over
items where a valid normalized answer can be extracted; in the final runs,
this is 621/621 items for all configurations.

\subsection{Overall Logical Performance}

Table~\ref{tab:overall-accuracy} summarizes performance. All evaluated models
achieve 91--94\% per-item accuracy, yet dialogue-level coherence is
substantially lower.

\begin{table}[t]
\centering
\small
\caption{Overall performance on \CBL.
Acc is per-item accuracy; DlgAcc is the fraction of dialogues where all
turns are correct.}
\label{tab:overall-accuracy}
\begin{tabular}{lccc}
\toprule
Model & Mode & Acc (\%) & DlgAcc (\%) \\
\midrule
GPT-4              & Zeroshot & 94.0 & 69.4 \\
GPT-4              & CoT      & 88.2 & 53.1 \\
Claude~3.5 Sonnet  & Zeroshot & 91.6 & 67.3 \\
Gemini~2.5 Flash   & Zeroshot & 91.9 & 71.4 \\
LLaMA-3 70B        & Zeroshot & 91.6 & 75.5 \\
LLaMA-3 70B        & CoT      & 89.5 & 65.3 \\
\bottomrule
\end{tabular}
\end{table}

\paragraph{Why DlgAcc is much lower than Acc.}
Dialogue accuracy compounds errors: for a dialogue of length $T$, even an
independent per-turn accuracy $p$ yields an ``all-correct'' probability of
approximately $p^T$. With $p \approx 0.92$ and $T \in [3,6]$, this ranges
from $0.92^3 \approx 0.78$ to $0.92^6 \approx 0.61$, matching the observed
DlgAcc range (53--76\%). In practice, dialogue errors are not independent:
models often drift in their interpretation of a system across
turns, causing clustered failures rather than isolated mistakes.

\paragraph{Local competence vs.\ global coherence.}
The high Acc values show that most single questions are solvable for
current LLMs. The lower DlgAcc values indicate that models struggle to
maintain a stable belief state across turns, even when the required beliefs
are simple boolean properties with deterministic implications.

\subsection{Task-Family Highlights and the ``0\% Compositional'' Gap}

Table~\ref{tab:task-champions} reports the best-performing model for key task
families. Models are strong on many atomic and multi-hop items but fail
completely on compositional questions.

\begin{table}[t]
\centering
\small
\caption{Task-specific best models on \CBL.
Accuracies are over items with a valid normalised answer.}
\label{tab:task-champions}
\begin{tabular}{lcc}
\toprule
Task family & Best model & Acc (\%) \\
\midrule
Atomic logical QA      & GPT-4 (Zeroshot) & 96.1 \\
Bias probes            & GPT-4 (Zeroshot) & 93.0 \\
Counterfactuals        & GPT-4 (Zeroshot) & 97.4 \\
Multi-hop reasoning    & GPT-4 (Zeroshot) & 97.1 \\
Multi-turn (per-turn)  & LLaMA-3 (Zeroshot) & 93.9 \\
Dialogue (dialogue-level) & LLaMA-3 (Zeroshot) & 75.5 \\
Compositional          & All models & 0.0 \\
\bottomrule
\end{tabular}
\end{table}

\paragraph{What makes compositional items qualitatively harder.}
Compositional questions deliberately combine multiple constraints, such as:
(i) chaining two or more implications under $\Phi$;
(ii) rejecting a misleading analogy cue;
(iii) applying a counterfactual change shown elsewhere in the dialogue; and
(iv) producing an answer consistent with prior turns. In practice, models
often succeed on each component in isolation but fail when the
components must be integrated into a single coherent decision.

\paragraph{A key observed failure mode: converse assumptions.}
Because $\Phi$ contains only one-way implications, compositional items often
punish the common LLM heuristic ``if A implies B, then B implies A.'' For
example, models may incorrectly infer $\Chaotic(s)$ from $\Sensitive(s)$ or
from $\PosLyap(s)$ even though the ontology does not assume those converses.
This is not a trick: it reflects real scientific reasoning discipline about
implication directionality.

\subsection{Consistency Diagnostics Beyond Accuracy}
\label{sec:diagnostics}

A major advantage of \CBL is that inconsistency is measurable.
Two models can achieve similar per-item accuracy while differing sharply in
how they fail. In particular, some errors are
local (a single wrong predicate assignment) while others represent
global constraint failures (e.g., asserting both $\Random(s)$ and
$\Deterministic(s)$, or implying chaos from a non-converse symptom).

These failures matter operationally: in downstream scientific workflows, a
locally incorrect answer may be caught by a user, but a globally
inconsistent belief state can produce confident yet mutually incompatible
explanations across time. This is precisely the failure mode that multi-turn
dialogues and axiom-violation tracking are designed to surface.

\subsection{Effect of Chain-of-Thought Prompting}
\label{sec:cot-analysis}

CoT is frequently assumed to improve reasoning~\cite{wei2022chainofthought},
but on \CBL it reduces both accuracy and dialogue-level coherence.

\begin{table}[t]
\centering
\small
\caption{CoT impact on overall and dialogue-level performance.}
\label{tab:cot-delta}
\begin{tabular}{lccc}
\toprule
Model & $\Delta$Acc (pp) & $\Delta$DlgAcc (pp) & Trend \\
\midrule
GPT-4 & $-5.8$ & $-16.3$ & worse \\
LLaMA-3 70B & $-2.1$ & $-10.2$ & worse \\
\bottomrule
\end{tabular}
\end{table}

\paragraph{Interpretation.}
CoT outputs are longer and appear more rigorous, but they can also introduce
extra unsupported assumptions (e.g., implicitly treating ``random-looking''
as ``random'') and amplify converse reasoning errors. CoT can therefore
increase explanatory confidence without improving logical
soundness. The larger drop in DlgAcc suggests that CoT especially harms
belief-state stability in multi-turn settings.

\section{Discussion}
\label{sec:discussion}

\subsection{What \CBL Reveals About LLM Reasoning}

\paragraph{The central finding: weak global constraint satisfaction.}
Across all evaluated models, single-question accuracy exceeds 90\%, yet
multi-turn coherence is far weaker. This indicates that current LLMs can
produce correct local judgments but struggle to maintain a consistent global
world model, even when the world model is as simple as a set of boolean
predicates linked by Horn-style implications.

\paragraph{Why scientific domains make this failure more obvious.}
In everyday QA, inconsistent answers can be rationalized away as ambiguity.
In dynamical systems, however, core relationships are crisp:
``random $\Rightarrow$ non-deterministic'' and
``chaotic $\Rightarrow$ deterministic and not random'' are definitional in
this ontology~\cite{strogatz2015nonlinear,ott2002chaos,devaney1989intro}.
\CBL turns these into explicit constraints that models must respect,
revealing when a model's internal reminders are not globally stable.

\subsection{How \CBL\ Can Support Neuro-Symbolic and Tool-Augmented Systems}

The structure of \CBL makes it a natural evaluation target for hybrid
approaches:

\begin{itemize}
  \item \textbf{Symbolic verifier / constraint checker.}
    An LLM can propose answers while a symbolic module checks whether the
    answers violate $\Phi$, triggering self-correction.
  \item \textbf{Logic-guided decoding.}
    Outputs can be constrained so that answers cannot jointly violate implications
    already committed within a dialogue.
  \item \textbf{Training with consistency losses.}
    Fine-tuning can penalize axiom violations across turns, not only wrong
    final labels.
\end{itemize}

Because \CBL uses a compact and explicit ontology, these methods are
implementable without full theorem proving.

\subsection{Limitations and Scope}

\CBL deliberately emphasizes logical structure over numerical
simulation. This makes the benchmark lightweight and auditable, but it also
means that each named system is treated as having a canonical regime label.
Future releases can extend the benchmark with (i) multiple parameter regimes
per system, (ii) items that require extracting regime information from
equations or parameter values, and (iii) hybrid items that combine symbolic
logic with tool-verified numeric diagnostics (e.g., Lyapunov estimation).

\section{Conclusion}
\label{sec:conclusion}

This paper introduces \CBL, a benchmark for diagnosing logical and symbolic reasoning
failures in LLMs within the domain of chaotic dynamical systems. By grounding
all questions in an explicit \FOL\ ontology, \CBL measures not only per-item
accuracy but also implication consistency and multi-turn coherence.
Experiments show that frontier and open-source LLMs achieve strong local
performance (91--94\% per-item accuracy) yet remain globally brittle:
dialogue-level accuracy is substantially lower and all tested models fail on
compositional items (0\%).
\CBL supports progress toward scientific AI systems that maintain
globally consistent beliefs, resist domain misconceptions, and integrate
symbolic structure with fluent language understanding.

\bibliography{references}

\end{document}